# The Dynamical Principles of Storytelling

Doxas[1,2], I., J. Meiss[3], S. Bottone[1], T. Strelich[4,5], A. Plummer[5,6], A. Breland[5,7], S. Dennis[8,9], K. Garvin-Doxas[9,10], M. Klymkowsky[3]

[1]Northrop Grumman Corporation
[2]Some work performed at the University of Colorado, Boulder
[3]University of Colorado, Boulder
[4]Fusion Constructive LLC
[5]Work performed at Northop Grumman Corporation
[6]Current Address JP Morgan
[7]Current address, GALT Aerospace
[8]University of Melbourne
[9]Work performed at the University of Colorado, Boulder
[10]Boulder Internet Technologies

When considering the opening part of 1800 short stories, we find that the first dozen paragraphs of the average narrative follow an action principle as defined in (3). When the order of the paragraphs is shuffled, the average no longer exhibits this property. The findings show that there is a preferential direction we take in semantic space when starting a story, possibly related to a common Western storytelling tradition as implied by Aristotle in *Poetics*.

## Introduction

Dynamical theory has long been applied to different aspects of human language. Reference (1) for example describes the dynamics involved in the way language changes over generations, and (2) describes the dynamics of the emergence of grammatical rules within an individual's mind. Recently, narrative has also been shown to be a dynamical system, and the trajectory of the average narrative between two points in semantic space shown to follow an action principle (3).

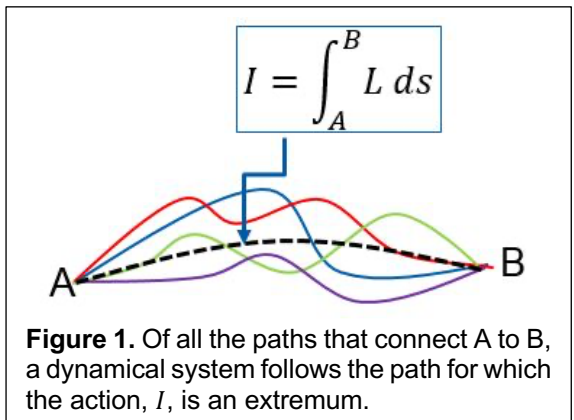

**Figure 1.** Of all the paths that connect A to B, a dynamical system follows the path for which the action, $I$, is an extremum.

Figure 1 illustrates the concept of the average narrative. As we embed paragraphs in a high dimensional semantic space, using standard tools like doc2vec (4, 5) or Latent Semantic Analysis (6), a narrative can be described by the trajectory that its successive paragraphs describe in that space. Given a starting and an ending paragraph, represented by points A and B in Figure 1, there are many comprehensible narratives that start with A and end with B. Reference (3) constructed a sample of these narratives in such a way as to have a readily computable mean, and showed that the mean path represents an extremum of the action integral

$$I = \int_A^B L \, ds \qquad [1]$$

where L is the Lagrangian of the system, and *s* is a parameter along the path, usually time. Although the strict definition of the classical path in mechanics only requires that the action be an extremum,

arXiv:2309.07797

it can be a minimum, and for narratives we find that a principle of *least* action holds.

Defining the Lagrangian of non-physical systems like narratives is not an obvious exercise, but if we follow a formalistic approach, and make some simplifying assumptions (e.g. field-free motion and discrete time; see (3)), we find that the action integral is given by the sum of the distances between the successive mean paragraphs in the embedding space,

$$I = \alpha \sum_{j=2}^{n} \left( \langle \vec{P_j} \rangle - \langle \vec{P_{j-1}} \rangle \right)^2 \quad [2]$$

where $n$ is the number of paragraphs in the path and

$$\langle \vec{P_j} \rangle = \frac{1}{N} \sum_{i=1}^{N} \vec{P_{ij}} \quad j = 1, \ldots, n \quad [3]$$

is the $j$th paragraph on the mean path, given by the average of the $j$th paragraphs of each of the N short story narratives. $\vec{P_{ij}}$ is the $j$th paragraph of the $i$th story ($1 \leq i \leq N, 1 \leq j \leq n$), and α is a proportionality constant.

In the present work we extend the application of the action principle to paths that are not anchored by fixed points, and show that, for a wide range of storytelling, there is a common starting part of narrative that is correctly ordered by an action principle.

**Methods and Results**

We start by embedding the first $n = 50$ paragraphs of each of $N = 1800$ short story narratives into a high dimensional semantic space. The narratives are a random and non-complete selection of short stories from Project Gutenberg, and represent a large cross-section of genres, including narratives that were not originally written in English. All stories have more than 50 paragraphs, but we look only at the first 50 paragraphs of each.

There are several methods for embedding text in a semantic space. We follow (3) and use doc2vec (4, 5) and Latent Semantic Analysis (6). For doc2vec we use one of the standard packages (gensim) without stop words. LSA weighs words using the information entropy; the word-document matrix, $w$, whose elements $w_{kl}$ give the number of times that word $l$ appears in document $k$, is weighted by

$$S_l = 1 + \frac{\sum_{k=1}^{M} Q_{kl} ln(Q_{kl})}{ln(M)} \quad [4]$$

to give the weighted word-document matrix

$$W_{kl} = S_l \, ln(w_{kl} + 1) \quad [5]$$

where

$$Q_{kl} = \frac{w_{kl}}{\sum_{k=1}^{M} w_{kl}} \quad [6]$$

is the probability of finding word $l$ in document $k$, and $M$ is the number of documents in the entire corpus. The matrix $W$ is approximated by keeping only a small number of singular values in its Singular Value Decomposition (SVD), thus drastically reducing the dimensionality of the space.

We use the largest 300 singular values (dimensions) for this work, but we have run tests with up to 500. We have also run doc2vec with dimensionalities ranging between 100 and 300, and find that the results vary little in this range. Results are also similar between methods; this is consistent with reference (7) which used, in addition to LSA, Non-negative Matrix Factorization (8), and the Topics Model (9). It was found that the variability between methods is comparable to the variability obtained by varying the number of narratives, $N$, for the average in [3].

After finding the document vectors, we compute the average path Since we are interested only in the start of the stories, and are looking at the first 50 paragraphs irrespective of the total length of each story, we can readily compute the average of each paragraph using [3]. The average path is then given by the sequence of points

$$\{\langle \vec{P}_1 \rangle, \langle \vec{P}_2 \rangle, \cdots, \langle \vec{P}_{50} \rangle\} \qquad [7]$$

We follow (3) and compute the Minimum Spanning Tree (MST) and the Acyclic Traveling Salesman Problem (A-TSP) for the sequence [7]. The A-TSP modifies the classic TSP, by allowing the "salesmen" to start and end their routes at different "cities" (while still visiting all the cities on the list exactly once). The MST finds the connections that minimize the sum of all distances when all points are connected once—in a tree—but not necessarily in a line. Thus, the MST does not include the concept of a sequence; it is a purely geometrical object.

Figure-2 shows the MST results, and Table-1 the A-TSP results, for the average path of the first 50 paragraphs using doc2vec embedding. We see that the A-TSP gives the correct sequence for the first 13 paragraphs. Furthermore, the MST shows that the configuration that minimizes the action (eq. 1) includes a linear, correctly ordered sequence of 14 paragraphs at the start. As a check, and to show that the ordering of the mean paragraphs is a property of the correct storytelling ordering of the paragraphs of each individual narrative, and not a property of the bag-of-paragraphs, we recompute the MST and A-TSP for the same narratives, but with the paragraph order of each narrative shuffled. In this case the mean paragraphs become

$$\langle \vec{P}_j \rangle = \frac{1}{N} \sum_{i=1}^{N} \vec{P}_{i\pi(i)_j} \qquad [8]$$

where $\pi(i)_j$ is the $j$th component of a random permutation of the paragraphs of narrative $i$, where $i = 1, \ldots, N$. Each $\langle \vec{P}_j \rangle$ is therefore still the average of N paragraphs, one from each narrative, but no longer the jth one from each narrative. Figure 2(b) shows that when the paragraphs of each narrative are shuffled, the mean paragraphs are no longer correctly ordered. The shuffling also disrupts the order of the solution to the A-TSP as shown in Table 1.

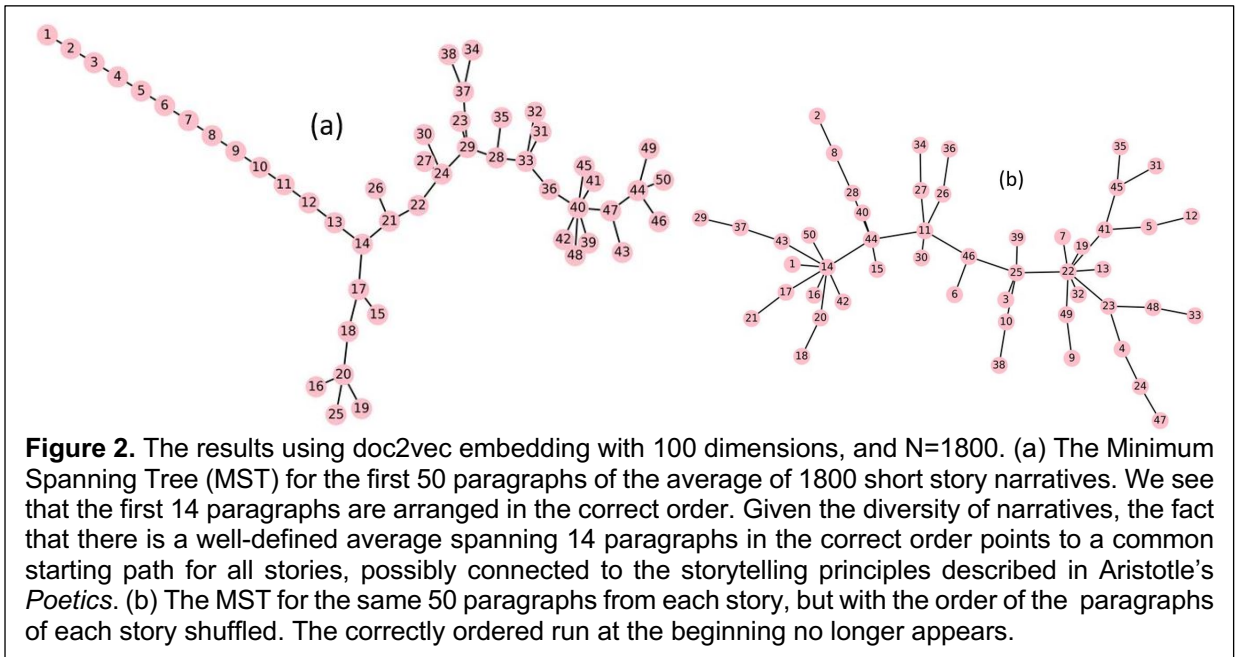

**Figure 2.** The results using doc2vec embedding with 100 dimensions, and N=1800. (a) The Minimum Spanning Tree (MST) for the first 50 paragraphs of the average of 1800 short story narratives. We see that the first 14 paragraphs are arranged in the correct order. Given the diversity of narratives, the fact that there is a well-defined average spanning 14 paragraphs in the correct order points to a common starting path for all stories, possibly connected to the storytelling principles described in Aristotle's *Poetics*. (b) The MST for the same 50 paragraphs from each story, but with the order of the paragraphs of each story shuffled. The correctly ordered run at the beginning no longer appears.

| Paragraphs in the correct storytelling order, doc2vec, N=1800 |
|---|
| 1 \| 2 \| 3 \| 4 \| 5 \| 6 \| 7 \| 8 \| 9 \| 10 \| 11 \| 12 \| 13 \| 15 \| 16 \| 17 \| 18 \| 20 \| 25 \| 14 \| 21 \| 22 \| 26 \| 30 \| 19 |
| 24 \| 27 \| 34 \| 38 \| 37 \| 35 \| 28 \| 29 \| 23 \| 32 \| 33 \| 31 \| 36 \| 42 \| 39 \| 48 \| 45 \| 43 \| 47 \| 40 \| 41 \| 49 \| 46 \| 44 \| 50 |
| Paragraphs shuffled, doc2vec, N=1800 |
| 1 \| 35 \| 16 \| 10 \| 29 \| 22 \| 19 \| 42 \| 5 \| 11 \| 45 \| 43 \| 9 \| 26 \| 32 \| 12 \| 27 \| 39 \| 18 \| 13 \| 7 \| 4 \| 41 \| 14 \| 17 |
| 31 \| 40 \| 46 \| 47 \| 44 \| 8 \| 33 \| 6 \| 21 \| 37 \| 23 \| 2 \| 15 \| 28 \| 36 \| 20 \| 49 \| 25 \| 38 \| 3 \| 24 \| 48 \| 34 \| 30 \| 50 |

**Table-1**. The order of the average paragraphs of the 1800 narratives, as given by the Acyclic Traveling Salesman Problem (A-TSP) which allows for the initial and final points to be different. We see that when the paragraphs of each narrative are in the correct order, the sequence that minimizes the action gives a correctly sequenced run at the start. We highlight the run for ease of reference. When the paragraphs of each narrative are shuffled, the minimum-action sequence no longer has the starting mean paragraphs in the correct order. The different color for paragraphs 15-18 is meant to denote the fact that 14 is missing from the sequence. But for that omission, the first eighteen paragraphs would have been correctly ordered.

The A-TSP is, of course, a more direct computation of the action principle, since it gives the sequence that minimizes the sum of successive distances (cf. eq. 2). Nevertheless, the MST provides more telling evidence of adherence to an action principle, by asking the more general question of finding the geometric arrangement that minimizes the sum of all distances. The fact that, at least for the first 14 paragraphs, that general arrangement is a linear path with the mean paragraphs in the correct order, points to a preferential direction and evolution of the start of narrative.

As a consistency check, we repeat the computation of the MST and A-TSP in two different ways: 1) by using LSA to embed the paragraphs in a 300 dimensional sematic space, and 2) computing the mean with a smaller random sample of $N = 500$ narratives instead of $N = 1800$ in [3]. In both cases we also recompute the metrics with the shuffled paragraphs for comparison.

Figure 3 and Table 2 show the results for the LSA case, and Figure 4 and Table 3 the results for the $N = 500$ case. We see that the basic result of a correctly ordered run at the start of the narrative is robust to using different embedding methods.

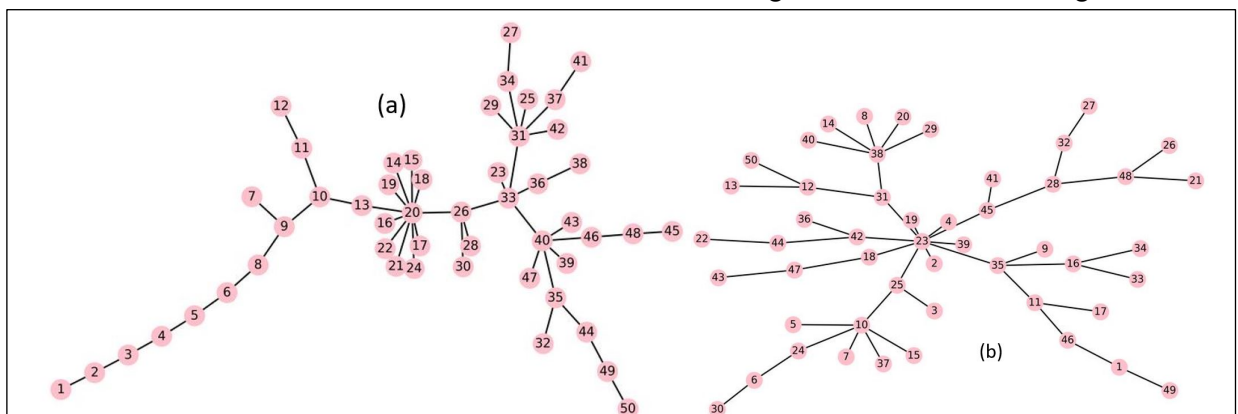

**Figure 3.** The MST results using LSA embedding with 300 dimensions, and N=1800. (a) The MST for the mean paragraphs with the paragraphs of each narrative in the correct order. Although the result is less striking than when using doc2vec for the embedding, there is clearly a correctly ordered run at the beginning. (b) The MST for the mean paragraphs, when the order of the paragraphs of each narrative is shuffled. The correctly ordered run at the beginning no longer appears.

| Paragraphs in the correct storytelling order, LSA, N=1800 |||||||||||||||||||||||||
|---|---|---|---|---|---|---|---|---|---|---|---|---|---|---|---|---|---|---|---|---|---|---|---|---|
| 1 | 2 | 3 | 4 | 5 | 6 | 8 | 7 | 9 | 10 | 11 | 12 | 13 | 16 | 15 | 17 | 14 | 21 | 25 | 18 | 20 | 24 | 22 | 19 | 28 |
| 26 | 30 | 29 | 31 | 37 | 41 | 34 | 27 | 43 | 39 | 40 | 35 | 32 | 23 | 33 | 36 | 38 | 42 | 44 | 49 | 45 | 48 | 46 | 47 | 50 |
| Paragraphs shuffled, LSA, N=1800 |||||||||||||||||||||||||
| 1 | 13 | 18 | 7 | 16 | 38 | 17 | 35 | 15 | 4 | 8 | 12 | 31 | 5 | 36 | 20 | 24 | 14 | 26 | 40 | 48 | 41 | 25 | 45 | 10 |
| 42 | 46 | 32 | 28 | 37 | 49 | 47 | 34 | 21 | 9 | 6 | 11 | 3 | 29 | 30 | 43 | 22 | 33 | 39 | 19 | 23 | 2 | 27 | 44 | 50 |

**Table-2**. Same as Table 1, but using LSA embedding. The results for the correctly ordered paragraphs are similar to those obtained with doc2vec embedding. The correctly ordered run at the start again disappears when the paragraphs of each narrative are shuffled. The different color for the paragraph sequence 9-13 denotes the fact that there is an inversion 7 ↔ 8 preceding it. But for that inversion, the first thirteen paragraphs would have all been correctly ordered.

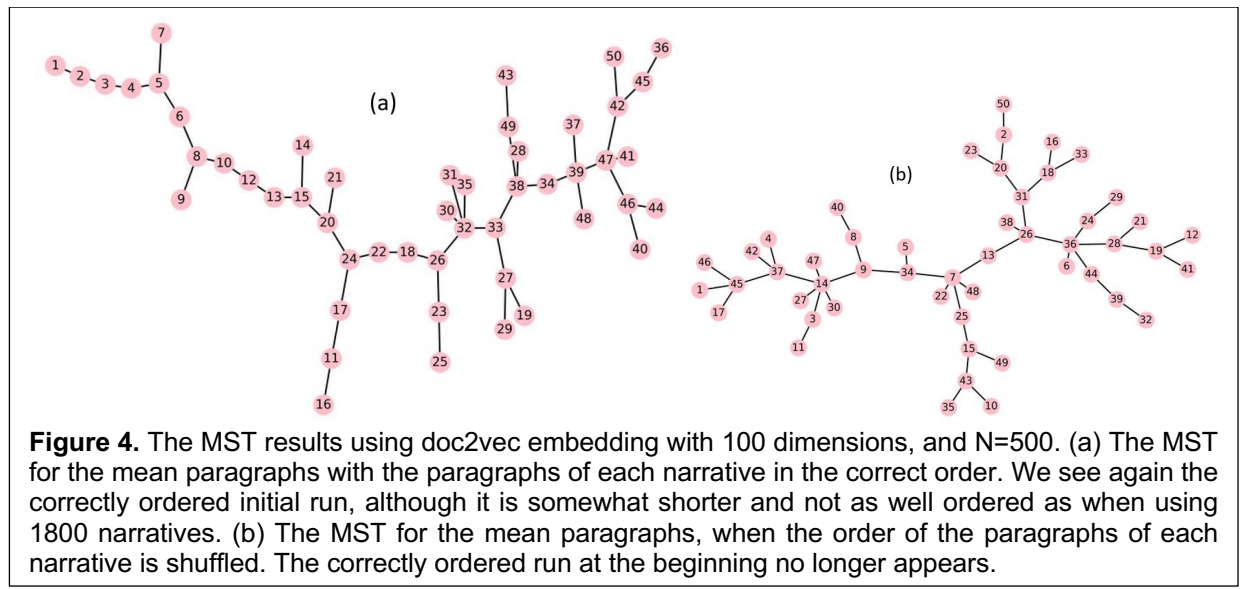

**Figure 4.** The MST results using doc2vec embedding with 100 dimensions, and N=500. (a) The MST for the mean paragraphs with the paragraphs of each narrative in the correct order. We see again the correctly ordered initial run, although it is somewhat shorter and not as well ordered as when using 1800 narratives. (b) The MST for the mean paragraphs, when the order of the paragraphs of each narrative is shuffled. The correctly ordered run at the beginning no longer appears.

| Paragraphs in the correct storytelling order, doc2vec, N=500 |||||||||||||||||||||||||
|---|---|---|---|---|---|---|---|---|---|---|---|---|---|---|---|---|---|---|---|---|---|---|---|---|
| 1 | 2 | 3 | 4 | 5 | 6 | 7 | 8 | 9 | 10 | 12 | 13 | 15 | 14 | 20 | 21 | 24 | 17 | 11 | 16 | 22 | 18 | 23 | 25 | 19 |
| 27 | 29 | 36 | 45 | 48 | 39 | 35 | 32 | 31 | 33 | 30 | 26 | 28 | 38 | 34 | 37 | 43 | 49 | 44 | 46 | 40 | 41 | 47 | 42 | 50 |
| Paragraphs shuffled, doc2vec, N=500 |||||||||||||||||||||||||
| 1 | 4 | 7 | 11 | 44 | 40 | 23 | 30 | 31 | 24 | 16 | 41 | 21 | 5 | 34 | 25 | 29 | 19 | 46 | 17 | 12 | 14 | 6 | 45 | 20 |
| 48 | 10 | 8 | 18 | 22 | 2 | 35 | 15 | 32 | 42 | 27 | 38 | 33 | 49 | 9 | 37 | 36 | 13 | 28 | 47 | 43 | 26 | 3 | 39 | 50 |

**Table-3**. Same as Table 1, but using a random selection of only 500 out of the 1800 narratives. There is clearly a correctly ordered initial run, although shorter than the N=1800 case. When the paragraphs of the narratives are shuffled, the run is clearly not present.

This result is consistent with previous work on narratives (3, 7, 10). We also see that reducing the number of narratives in the average makes the average sequence less well defined, as would be expected.

**Conclusions**

When describing style writers, Science Fiction writer James Gunn (11) said that the only way style writers can get away with always using the same style, is by telling the same story over and over again. Literature is no stranger to telling the same tale in many different ways. Indeed, Aristotle's *Poetics*, by formulating the rules for a well-told story, laid down the rules as he saw them for the structure of *all* stories. For the beginning of a story in particular, the implication is that there is a universal way to start a story in order that it evolve effectively, and our results confirm that there is a well-defined trajectory around which the beginnings of stories evolve. Although individual narratives are, of course, very different, the way opening paragraphs follow one after the other has a well-defined common characteristic, namely that the average of all opening sequences minimizes the action [1]. It is a subject of further research whether this principle applies to all human storytelling, and whether other cultures' storytelling traditions evolve according to similar or different principles.

## Acknowledgments

ID and SD wish to acknowledge many stimulating discussions with the late prof. Walter Kintsch. This work was partly supported by a grant from the National Science Foundation.